\documentclass[11pt,a4paper]{article}
\usepackage{times,latexsym}
\usepackage{url}
\usepackage[T1]{fontenc}

\usepackage[acceptedWithA]{tacl2018v2}
\usepackage{amsmath}
\usepackage{graphicx}
\usepackage{booktabs}
\usepackage{amssymb}
\usepackage{transparent}
\usepackage{multirow}
\usepackage{tikz}
\usepackage{cleveref}
\usepackage{pifont}
\usepackage[super]{nth}
\usepackage[shortlabels]{enumitem}
\usepackage{quoting}
\crefformat{section}{\S#2#1#3} % see manual of cleveref, section 8.2.1
\crefformat{subsection}{\S#2#1#3}
\crefformat{subsubsection}{\S#2#1#3}
\newcommand{\cmark}{\ding{51}}%
\newcommand{\xmark}{\ding{55}}%

\usepackage{xspace,mfirstuc,tabulary}

\newif\iftaclinstructions
\taclinstructionsfalse % AUTHORS: do NOT set this to true
\iftaclinstructions
\renewcommand{\confidential}{}
\renewcommand{\anonsubtext}{(No author info supplied here, for consistency with
TACL-submission anonymization requirements)}
\newcommand{\instr}
\fi

\iftaclpubformat % this "if" is set by the choice of options

\else

\fi

\title{Maintaining Common Ground in Dynamic Environments}

\author{Takuma Udagawa$^1$ \: and \: Akiko Aizawa$^{1,2}$ \\  
	The University of Tokyo, Tokyo, Japan$^1$ \\
	National Institute of Informatics, Tokyo, Japan$^2$\\
	\texttt{\{takuma\_udagawa,aizawa\}@nii.ac.jp}}

\date{}

\begin{document}
\maketitle
\begin{abstract}
Common grounding is the process of creating and maintaining mutual understandings, which is a critical aspect of sophisticated human communication. While various task settings have been proposed in existing literature, they mostly focus on creating common ground under static context and ignore the aspect of \textit{maintaining} them overtime under \textit{dynamic} context. In this work, we propose a novel task setting to study the ability of both creating and maintaining common ground in dynamic environments. Based on our minimal task formulation, we collected a large-scale dataset of 5,617 dialogues to enable fine-grained evaluation and analysis of various dialogue systems. Through our dataset analyses, we highlight novel challenges introduced in our setting, such as the usage of complex \textit{spatio-temporal expressions} to create and maintain common ground. Finally, we conduct extensive experiments to assess the capabilities of our baseline dialogue system and discuss future prospects of our research.
\end{abstract}

\section{Introduction}
\label{sec:introduction}

\begin{figure*}[tb!]
\centering
\includegraphics[width=\textwidth]{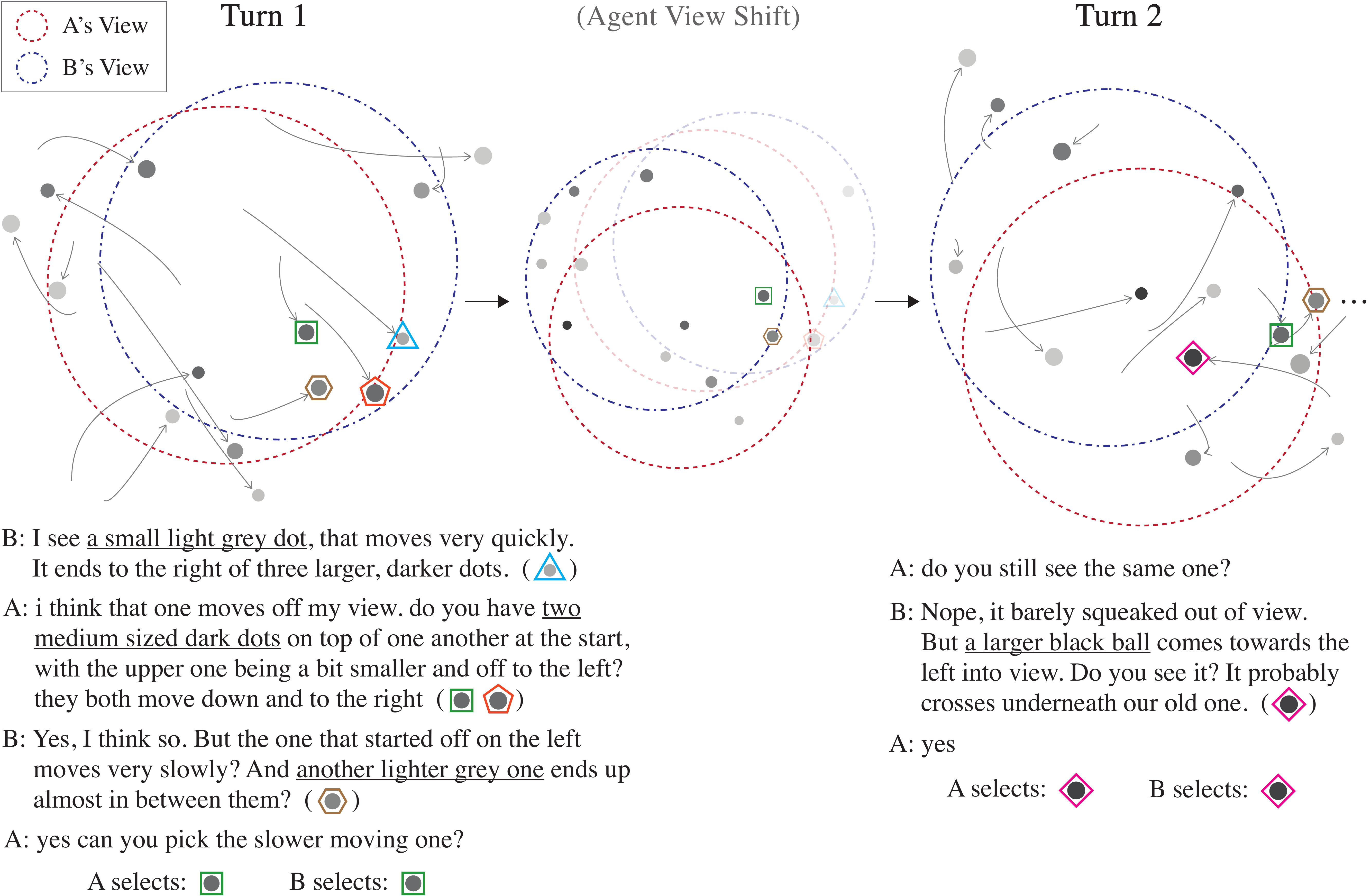}
\caption{Example dialogue of our sequential collaborative reference task (\cref{sec:task_formulation}). Each agent has a partial view of a 2-D plane with synthetic entities (grayscale dots of various sizes). \textit{During} each turn, the entities move randomly on the 2-D plane. \textit{At the end} of each turn, the agents communicate with each other to find and select one of the same, common entities. \textit{After} each turn (if the selections match), both agents' views shift randomly and the next turn begins. *\,Note that the colored polygons (indicating the referents of the underlined expressions) are shown for illustration purposes only and not visible to the agents nor provided in the current dataset.
}
\label{fig:first_example}
\end{figure*}

Common grounding is the process of creating, repairing and updating mutual understandings (i.e. \textit{common ground}), which is a critical aspect of sophisticated human communication \citep{clark1996using}. Humans can \textit{create} substantial common ground by expressing various information in natural language, which can be clarified or \textit{repaired} to resolve misunderstandings at essential levels of detail. Furthermore, as the situation changes and relevant information gets outdated, humans can \textit{update} their common ground accordingly by discarding old information and acquiring new ones. Such ability plays a vital role in sustaining collaborative relationships and adapting to emerging problems in nonstationary, real-world environments.

However, despite the wide variety of tasks proposed in existing literature \citep{fang2015embodied,zarriess-etal-2016-pentoref,de2017guesswhat,udagawa2019natural,haber-etal-2019-photobook}, they mostly focus on creating common ground under \textit{static} (time-invariant) context and ignore their \textit{dynamic} aspects. While some recent dialogue tasks deal with dynamic information, they often lack suitable evaluation metrics \citep{pasunuru-bansal-2018-game}, \textit{context updates} in the course of the dialogue \citep{alamri2019audio} or diverse dynamics of the environment itself \citep{de2018talk,suhr-etal-2019-executing,narayan-chen-etal-2019-collaborative,thomason:corl19,moon-etal-2020-situated}. Therefore, it remains unclear how well existing dialogue systems can adapt to the diversely changing situations through advanced common grounding (\cref{sec:related_work}).

To address this problem, we propose a novel dialogue task based on three design choices (\cref{sec:task_formulation}):

First, we formulate a novel \textit{sequential collaborative reference task} as a temporal generalization of the collaborative reference task proposed in \citet{he2017learning} and \citet{udagawa2019natural}. In our formulation, the goal of the agents is generalized to track and select the common entity \textit{at multiple timesteps}, while the agents' observations change dynamically between each timestep. This setting requires both \textit{creation} and \textit{maintenance} of common ground, whilst enabling clear evaluation based on the length of successful timesteps.

Secondly, we focus on synthesizing the \textit{entity movements}, as popularized in the recent video understanding benchmarks \citep{girdhar2020cater,yi2020clevrer,bakhtin2019phyre}. By leveraging such synthetic dynamics, we can minimize undesirable biases, maximize diversity and enable fully controlled evaluation and analysis.

Finally, we build upon OneCommon Corpus \citep{udagawa2019natural} to introduce natural difficulty of common grounding with minimal task complexity. To be specific, we represent entity attributes and their temporal dynamics based on \textit{continuous} real values to introduce high ambiguity and uncertainty. In addition, we consider a \textit{partially-observable} setting where each agent only has a partial view of the environment, which introduces various misunderstandings and partial understandings that need to be resolved.

Based on this task design, we collected a large-scale dataset of 5,617 dialogues (including over 65K utterances) through careful crowdsourcing on Amazon Mechanical Turk (\cref{sec:dataset_collection}).

We show an exemplary dialogue of our task in Figure \ref{fig:first_example}. Since the environment is dynamic, humans rely on various \textit{spatio-temporal expressions} to express entity states at different timesteps (``\underline{started off} on the left'', ``\underline{ends} to the right'') or how they changed dynamically (``moves very quickly'', ``come towards the left'') to create common ground. Furthermore, in later turns, humans often leverage their \textit{previous common ground} (``still see the same one?'', ``crosses underneath our old one'') to update their common ground more reliably and efficiently. We conduct detailed analyses of the dataset to study such strategies in \cref{sec:dataset_analysis}.

In our experiments (\cref{sec:experiments}), we train a neural-based dialogue system based on \citet{udagawa2020annotated}. Through our extensive evaluation and analysis, we assess the current model's strengths as well as important limitations and demonstrate huge room left for further improvement.

Overall, our main contributions are:

\begin{itemize}[topsep=0pt, itemsep=0pt, leftmargin=.2in, parsep=0pt]
    \item Proposal of a novel dialogue task to study common grounding in dynamic environments.
    \item Large-scale dataset of 5,617 dialogues to develop and test various data-driven models.\footnote{Our code and dataset are publicly available at \url{https://github.com/Alab-NII/dynamic-onecommon}.} 
    \item Detailed dataset analyses which highlight novel challenges introduced in our setting.
    \item Extensive evaluation and analysis of a simple yet strong baseline dialogue system.
\end{itemize}

\begin{table*}[th!]
\centering \scalebox{0.67}{
\setlength\tabcolsep{6pt}
\begin{tabular}{ccccccc}
\toprule[\heavyrulewidth]
\multirow{2}{*}[-0.8ex]{\textbf{Dataset}} & \multicolumn{3}{c}{\textbf{Environment (Context Type)}} & \raisebox{-2pt}{\textbf{Context}} & \raisebox{-2pt}{\textbf{Context}} & \raisebox{-2pt}{\textbf{Evaluation of}} \\
\cmidrule(r){2-4}
& \textbf{Continuous} & \textbf{Partially-Observable} & \textbf{Dynamic} & \raisebox{2pt}{\textbf{Update}} & \raisebox{2pt}{\textbf{Source}} & \raisebox{2pt}{\textbf{Common Grounding}} \\
\midrule
Twitch-FIFA \citep{pasunuru-bansal-2018-game} & \cmark & \xmark & \cmark & \cmark & Synthetic & N/A \\
AVSD \citep{alamri2019audio} & \cmark & \cmark & \cmark & \xmark & Real & Indirect \\
SIMMC \citep{moon-etal-2020-situated} & \cmark & \xmark & \xmark & \cmark & Synthetic+Real & Indirect \\
MutualFriends \citep{he2017learning} & \xmark & \cmark & \xmark & \xmark & Synthetic & Create \\
GuessWhat?! \citep{de2017guesswhat} & \cmark & \xmark & \xmark & \xmark & Real & Create \\
Photobook Dataset \citep{haber-etal-2019-photobook} & \cmark & \cmark & \xmark & \cmark & Real & Create \\
OneCommon \citep{udagawa2019natural} & \cmark & \cmark & \xmark & \xmark & Synthetic & Create \\
\midrule
\textbf{Dynamic-OneCommon (Ours)} & \cmark & \cmark & \cmark & \cmark & \textbf{Synthetic} & \textbf{Create+Maintain} \\
\bottomrule[\heavyrulewidth]
\end{tabular}
}
\caption{
Comparison with the major datasets. Environments are considered \textit{dynamic} if they involve rich, spontaneous dynamics and contexts to be \textit{updated} if new information is provided in the course of the dialogue.
}
\label{tab:dataset_comparison}
\end{table*}

\section{Related Work}
\label{sec:related_work}

The notion of common ground was originally introduced in \citet{lewis1969convention} and \citet{stalnaker1978assertion} and theoretically elaborated in fields such as psycholinguistics \citep{clark1991grounding,brennan2010two}. While formal approaches (rule/logic-based) exist to computationally model the process of common grounding \citep{traum1994computational,van2007dynamic,poesio2010completions}, capturing their full complexities in realistic, situated conversations remains a formidable problem.

From an empirical perspective, various dialogue tasks have been proposed to develop and evaluate data-driven models of common grounding. Most of the existing literature focuses on closed domain, goal-oriented settings to measure the ability both quantitatively and objectively \citep{fang2015embodied,zarriess-etal-2016-pentoref,de2017guesswhat}. Recent works, summarized as the \textit{grounded agreement games} in \citet{schlangen2019grounded}, introduce symmetric speaker roles to encourage more bilateral interaction. \citet{udagawa2019natural} also raise \textit{continuous} and \textit{partially-observable} context to be essential for requiring advanced common grounding (\cref{subsec:collaborative_reference_task}). Finally, \citet{haber-etal-2019-photobook} propose a multi-round image identification task, where different combinations of images are provided to each agent at every round. While this setting is useful for studying \textit{subsequent references} affected by the existing common ground \citep{Brennan1996ConceptualPA,takmaz-etal-2020-refer}, the observations in each round are static, temporarily independent images. Hence, all of these tasks focus on creating common ground under \textit{static} context and lack evaluation metrics for \textit{maintaining} common ground in dynamic environments.

We also note that some recent dialogue tasks require dealing with dynamic information, although common grounding usually takes place \textit{implicitly} and may be difficult to measure directly. For instance, \citet{alamri2019audio} proposed Q\&A based dialogues grounded in video contexts. However, the information given to each agent remains fixed throughout the dialogue, requiring \textit{creation} but minimal \textit{update} of common ground. Many recent works also focus on dialogues grounded in external environments \citep{de2018talk,suhr-etal-2019-executing,narayan-chen-etal-2019-collaborative,thomason:corl19,moon-etal-2020-situated}. These settings often involve dynamic change of the \textit{perspectives}, but they usually assume the environments themselves to be stationary and do not change spontaneously (without direct intervention). In contrast to these works, we introduce both \textit{context updates} in the course of the dialogue and \textit{diverse dynamics} of the external environment to require advanced common grounding.\footnote{While \citet{pasunuru-bansal-2018-game} collected live-stream dialogues grounded in soccer video games, the non-goal-oriented, unconstrained nature of their setting makes evaluation and analysis of common grounding very challenging.} We summarize our comparison with the major existing datasets in Table \ref{tab:dataset_comparison}.

Finally, our work is relevant to the emerging literature on spatio-temporal grounding in computer vision and NLP. This includes video QA \citep{lei-etal-2018-tvqa,yu2019activityqa,castro-etal-2020-lifeqa}, video object grounding \citep{ZhLoCoBMVC18,chen-etal-2019-weakly,Sadhu_2020_CVPR} and video captioning \citep{krishna2017dense}, all of which are essential subtasks in our dialogue. However, existing resources often contain exploitable biases and lack visual/linguistic diversity as well as reliable evaluation metrics (esp. in language generation) \citep{aafaq2019video}. It is also challenging to probe model behaviors without the controllability of the video contexts \citep{girdhar2020cater}. We have addressed such concerns based on our task design (\cref{subsec:sequential_collaborative_reference_task}) and expect our resource to be useful for promoting this line of research as well.

\section{Task Formulation}
\label{sec:task_formulation}

In this section, we review the collaborative reference task from OneCommon Corpus (OCC in short) and formulate our \textit{sequential} counterpart as its temporal generalization.

\subsection{Collaborative Reference Task}
\label{subsec:collaborative_reference_task}

Based on \citet{udagawa2019natural}, a \textit{collaborative reference task} is a multi-agent cooperative game with entities $E = \{e_1, e_2, ... , e_m\}$ and agents $A = \{a_1, a_2, ... , a_n\}$. Each agent $a_j \in A$ has an observation of entities $obs_j(E)$ and can exchange information with other agents in natural language. At the end of the game, each agent selects one of the observable entities, and the game is \emph{successful} if and only if all the agents selected the same entity.\footnote{In contrast to the typical reference tasks \citep{de2017guesswhat}, agent roles are \textit{symmetric} and they can agree upon any of the common entities (as long as it's the same).} This can be considered as a general framework for evaluating accurate \textit{mutual recognition} of a common entity, which is often a critical step in general common grounding.

\begin{figure}[ht!]
\centering
\includegraphics[width=0.99\columnwidth]{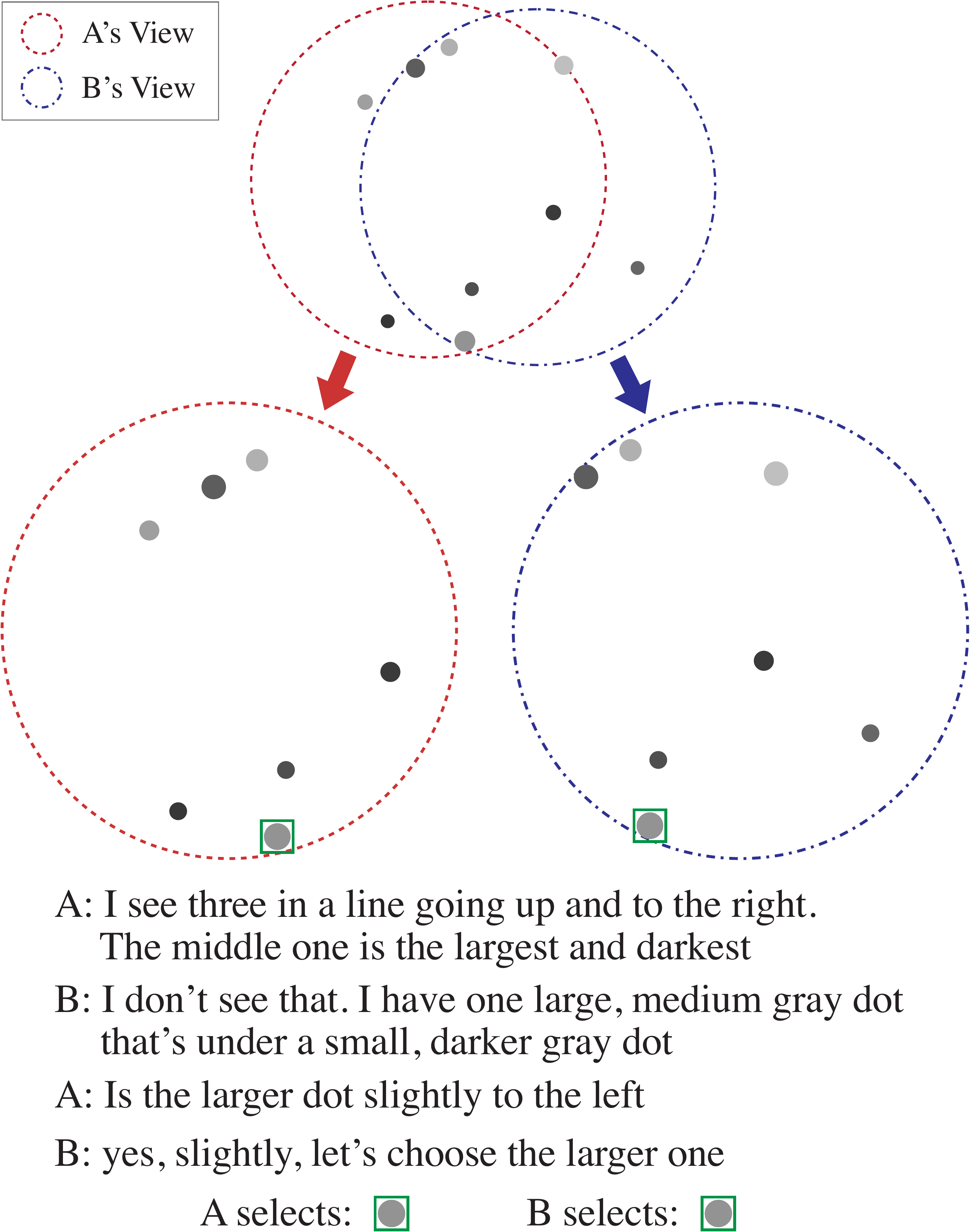}
\caption{Example dialogue from OneCommon Corpus (OCC). We can see that the human players are able to detect misunderstandings and make flexible clarifications to reduce ambiguity and uncertainty.
}
\label{fig:onecommon_example}
\end{figure}

One main feature of OCC is that they represented all entity attributes (color, size and location on a 2-D plane) based on \textit{continuous} real values. Unlike discrete/categorical attributes, this introduces high ambiguity and uncertainty to be expressed in symbolic natural language. In addition, they introduced \textit{partial-observability} where each agent only has a partial view of the 2-D plane, which requires collaborative resolution of various misunderstandings. We show an example of a successful dialogue from OCC in Figure \ref{fig:onecommon_example}.

However, this current task formulation assumes each observation to be \textit{static} and can only evaluate the ability of \textit{creating} common ground.

\subsection{Sequential Collaborative Reference Task}
\label{subsec:sequential_collaborative_reference_task}

To address this limitation, we generalize each observation to be \textit{dynamic} and collaborative reference to be \textit{sequential}. Specifically, each agent $a_j \in A$ now receives observation $obs_j(E, t)$ at each timestep $t \in [t_0, \infty)$, and the agents' goal is to communicate in natural language to select the same entity at multiple timesteps $t_1, t_2, ... \in (t_0, \infty)$.\footnote{We assume $t_{k -1} < t_k $ for all $k \in \mathbb{N}$.} At each selection timestep $t_k$ ($k \in \mathbb{N}$), $a_j$ must select one entity observable at $t_k$ but has all previous observations up to $t_k$, $\{obs_j(E, t) | t \in [t_0, t_k] \}$. The game ends when the selections no longer match at timestep $t_{k^\prime}$ ($k^\prime \in \mathbb{N}$): therefore, the success at $t_1$ measures the ability of \textit{creating} common ground, and the length of successful timesteps (LST) $k^\prime-1$ measures the ability of \textit{maintaining} them. This is a general framework for evaluating both creation and maintenance of mutual entity recognition in dynamic environments.

Based on this task formulation, we propose a minimal task setting extending OCC and incorporate dynamic change of the entity \textit{locations}.

\begin{figure}[th]
\centering
\includegraphics[width=0.75\columnwidth]{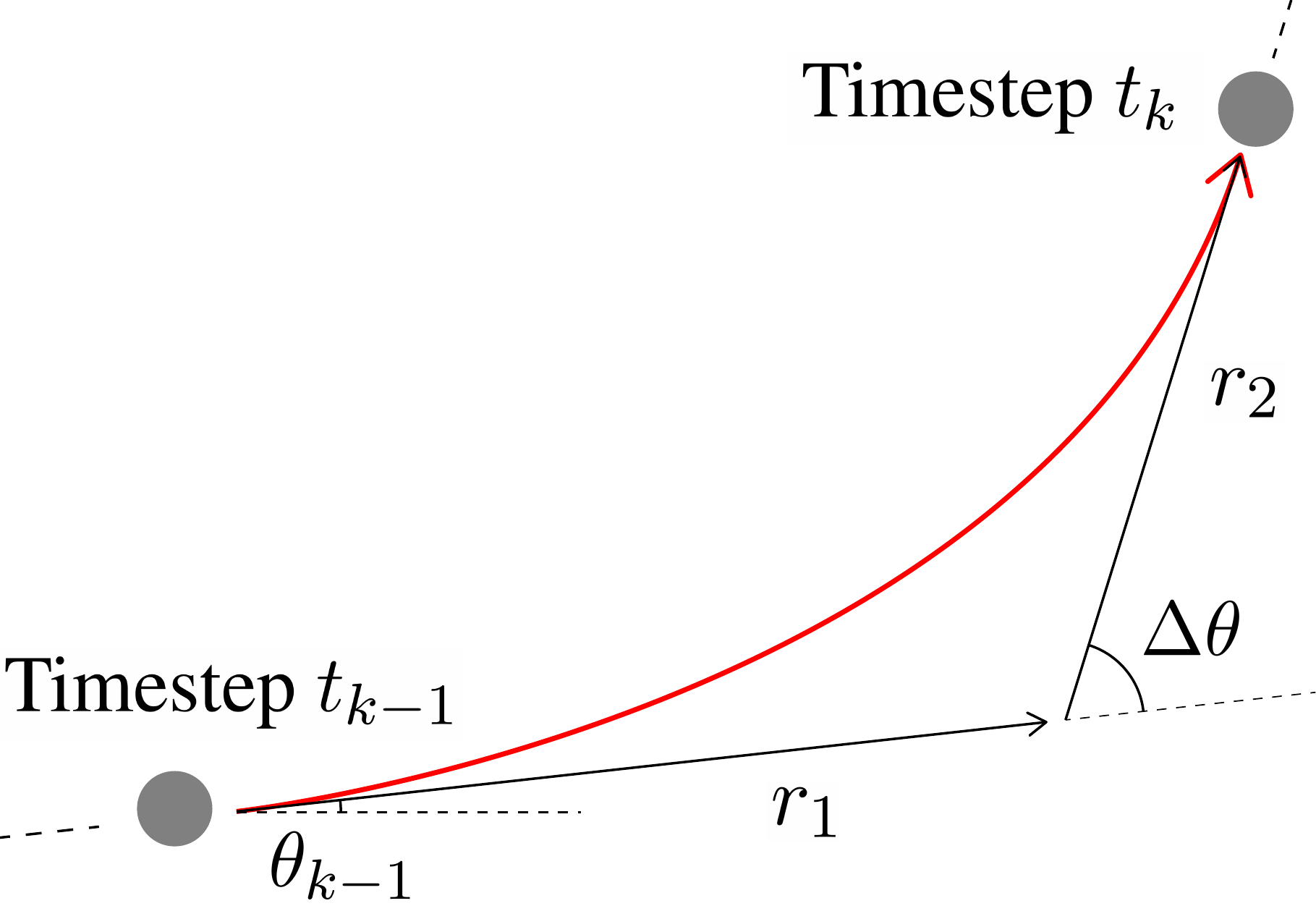}
\caption{Illustrated movement of each entity in turn $k$.
}
\label{fig:entity_movement}
\end{figure}

We refer to each time range $[t_{k-1}, t_k]$ as \textit{turn} $k$. During each turn, we change the location of each entity $e_i \in E$ based on a simple parameterized movement, where the \textit{trajectory} is determined by a quadratic B\'{e}zier curve \citep{bezier1974mathematical}.\footnote{Its \textit{speed} is proportional to the length of the trajectory.} See Figure \ref{fig:entity_movement} for an illustration, where $r_{1}$, $r_{2}$ are parameters of \textit{distance} and $\theta_{k-1}$, $\Delta \theta$ represent \textit{angles}. We sample $r_{1}$, $r_{2}$, $\Delta \theta$ from fixed uniform distributions each turn and update $\theta_{k}$ as $\theta_{k} \leftarrow \theta_{k-1} + \Delta \theta$ ($\theta_{0}$ is initialized randomly). This way, we can generate diverse, unbiased, coherent and fully controllable dynamics of the environment.

To enable fair comparison with OCC, we limit the number of agents to 2 and set the circular agent views to have the same diameter as OCC. At each selection timestep $t_k$, we ensure that each agent has 7 observable entities with only 4, 5 or 6 of them in common, which is also identical to OCC. Finally, we sample all entity attributes (color, size and initial location) from the same uniform distributions as OCC with minimal modifications.\footnote{To be specific, we set the minimum distance between entities (at $t_k$) and the possible range of entity size to be slightly different to avoid entity overlapping during movements.} Therefore, we expect the (distribution of) observations at $t_k$ to be similar and enable mostly fair comparison with OCC (in \cref{sec:dataset_analysis} and \cref{sec:experiments}).

To ensure task difficulty, we also shift the \textit{perspective} of each agent after each successful turn (see Figure \ref{fig:first_example}) so that the overlapping regions differ every turn. The same dot is prohibited from staying in common for over 3 consecutive selection timesteps, requiring frequent updates of common ground. Finally, we limit the maximum number of turns to 5 for practical purposes (hence the maximum LST is 5 in each game).

\section{Dataset Collection}
\label{sec:dataset_collection}

\begin{figure*}[th]
\centering
\includegraphics[width=0.97\textwidth]{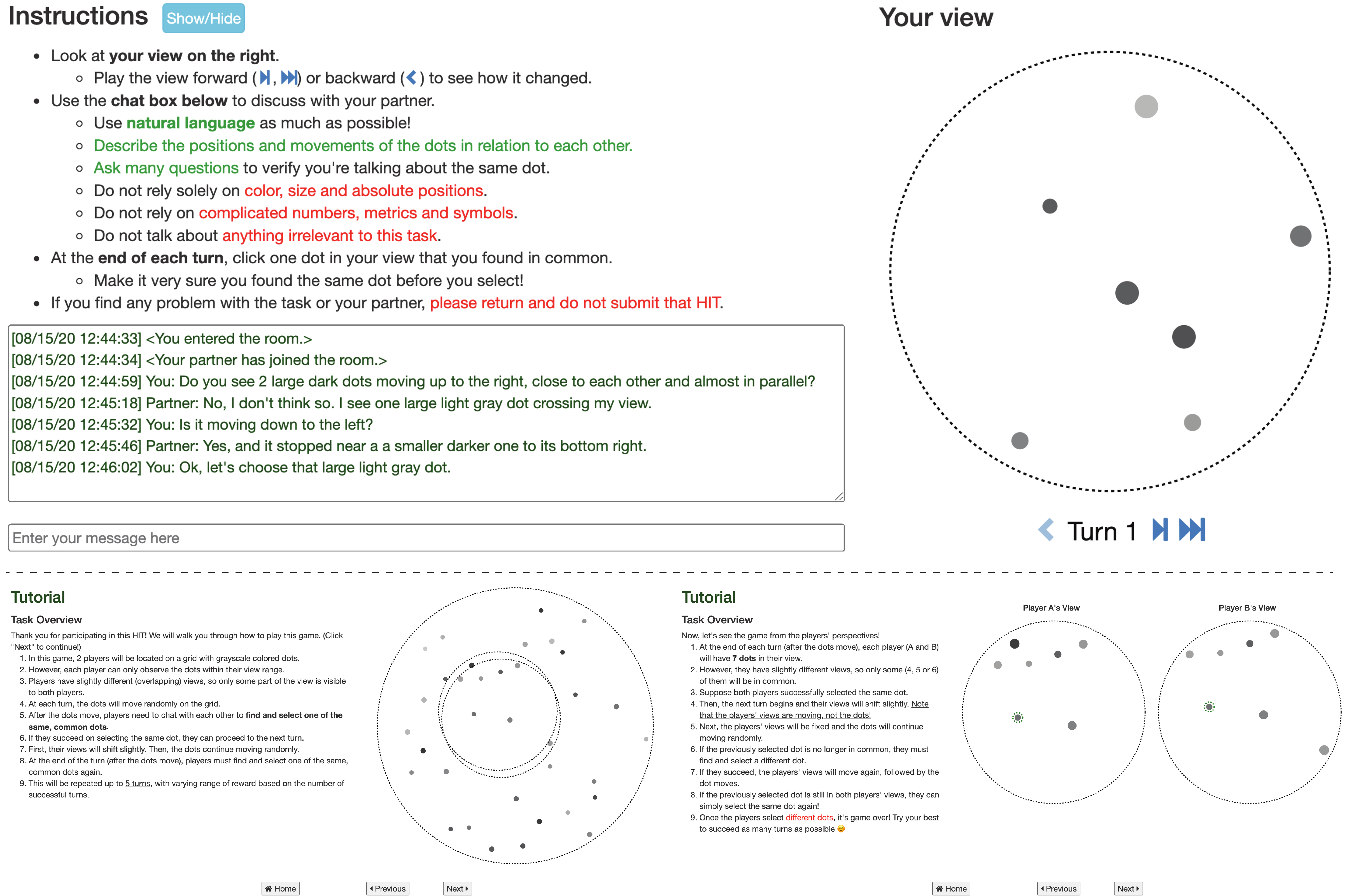}
\caption{(Top) Our dialogue interface. During the game, animations up to the current turn could be replayed anytime using the forward/backward buttons. (Bottom) Sample screenshots from our tutorial on the \textit{task setting}.
}
\label{fig:user_interfaces}
\end{figure*}

To collect large-scale, high-quality dialogues, we conducted careful crowdsourcing on Amazon Mechanical Turk. The web application is based on the CoCoA framework \citep{he2017learning}, and we used Scalable Vector Graphics (SVG) to animate entity movements and parallel shifts of the agent perspectives. Before working on our task, crowd workers were required to take a brief tutorial on the task setting, dialogue interface and instructions. Sample screenshots of our dialogue interface and tutorial are shown in Figure \ref{fig:user_interfaces}: note that animations up to the current turn could be replayed anytime for the ease of playing the game.\footnote{This also allows us to ignore the disadvantage of imperfect human memories in comparison to machines.}

To ensure worker quality, we required crowd workers to have more than 500 completed HITs and acceptance rates higher than 99\%. To encourage success, we rewarded \$0.25 for every successful turn plus additional bonuses for longer LST achieved (up to \$0.25 if LST $=$ 5). Finally, we manually reviewed all submitted works and excluded dialogues which clearly violated the instructions (e.g. relying on premature guessing or other ineffective strategies\footnote{Typical examples include strategies relying solely on color, size and absolute positions in the agent's view.}). We did not exclude dialogues based on task failures (even if LST $=$ 0), as long as they were based on valid strategies.

To solicit linguistic/strategic variety, we generally used a unique environment for each game. However, if the task was unsuccessful (i.e. LST $=$ 0), we allowed the environment to be reused in another game. This way, we can expect to eventually collect successful (LST $>$ 0) dialogues for the relatively difficult environments as well.

Overall, we collected 5,804 dialogues, and after the reviewing process, we were left with 5,617 qualified dialogues. We refer to this dataset as \textbf{Dynamic-OneCommon Corpus (D-OCC)}. Note that our dataset is currently in English, but the dataset collection procedure is language-agnostic and can be applied in any other languages.

\section{Dataset Analysis}
\label{sec:dataset_analysis}

Next, we conduct detailed analyses of the dataset to study human common grounding strategies under dynamic context. Whenever possible, we give comparative analyses with OCC to highlight the effect of dynamic factors introduced in D-OCC.

\subsection{Overall Statistics}
\label{sec:overall_statistics}

\begin{table}[th!]
\centering \scalebox{0.85}{
\setlength\tabcolsep{8pt}
\begin{tabular}{lcc}
\toprule[\heavyrulewidth]
Statistics & OCC & D-OCC \\
\midrule
Total dialogues & 6,760 & 5,617 \\
Uttrances per dialogue & 4.8 & 11.7 \\
Tokens per utterance & 12.4 & 10.3 \\
Duration per dialogue (minutes) & 2.1 & 5.7 \\
Unique workers & N/A & 462 \\
Avg. LST & - & 3.31 \\
Avg. completed turns & - & 3.77 \\
\midrule
Unique tokens & 3,621 & 3,895 \\
Occupancy of rare tokens (\%) & 1.4 & 1.0 \\
Overlap of all tokens (\%) & \multicolumn{2}{c}{29.4} \\
Overlap w/o rare tokens (\%) & \multicolumn{2}{c}{53.0} \\
\bottomrule[\heavyrulewidth]
\end{tabular}
}
\caption{
Statistics of OCC and D-OCC datasets.
}
\label{tab:onecommon_statistics}
\end{table}

First, we summarize the overall statistics of OCC and D-OCC in Table \ref{tab:onecommon_statistics}.

In total, OCC and D-OCC have a comparable number of dialogues. However, dialogues can be much longer in D-OCC, since collaborative reference is repeated multiple times. On average, utterance lengths are slightly shorter in D-OCC: this can be mostly attributed to the increased (relative) frequency of short utterances like acknowledgments and \textit{shortened} subsequent responses (e.g. ``same again?'' $=$ ``select the \underline{same} black dot \underline{again}?'').\footnote{In fact, utterances with less than 5 tokens were almost twice more frequent in D-OCC (33.8\%) than OCC (17.6\%).} Note that long, complex utterances are also common in our dataset, as seen in Figure \ref{fig:first_example}. Overall, we found 462 unique workers participated in D-OCC, which indicates reasonable diversity at the \textit{player} level as well.

In terms of LST, the overall average was 3.31 with over half (53.5\%) of the dialogues succeeding all 5 turns. This suggests that humans can solve the task reliably through sophisticated common grounding. After filtering dialogues with poor/careless workers (whose avg. LST $<$ 2), we observed a slight improvement up to 3.57. If we only focus on the top 10 workers (with at least 10 tasks completed), avg. LST was significantly higher reaching 4.24. These results indicate that (at least potentially) much higher human ceiling performance can be achieved. Note that if we include the last unsuccessful turn in 46.5\% of the dialogues, the average of all completed turns was slightly longer (3.77) in our dataset.

Finally, we found that both datasets have a relatively small vocabulary size as well as the occupancy of \textit{rare tokens} (used less than 10 times in the dataset).\footnote{Occupancy is computed based on the proportion of total frequencies (TF), i.e. \textit{TF of rare tokens} / \textit{TF of all tokens}.} This indicates minimal complexity at the \textit{lexical} level, as observed in \citet{udagawa2019natural}. We also found that the two datasets have a large vocabulary overlap, which is expected as D-OCC extends the setting of OCC.

\subsection{Spatio-Temporal Expressions}
\label{subsec:spatio_temporal_expressions}

\begin{table*}[th]
\centering \scalebox{0.83}{
\setlength\tabcolsep{6pt}
\newcommand{\intermidrule}{\cmidrule{2-3}} 
\begin{tabular}{llcc}
\toprule
\multicolumn{1}{c}{Reference} & \multicolumn{1}{c}{Examples} & Freq. & Cohen's $\kappa$ \\
\midrule
\multirow{4}{*}{Current State} &It's to the right of where the grey one \underline{ended up} for me \underline{after} moving up and left. & \multirow{4}{*}{23.8\%} & \multirow{4}{*}{0.91} \\
&\underline{Now} I have another triangle / Does it \underline{land} next to two smaller gray dots?& & \\
&Does it have a lighter one below and to the left \underline{when they stop}?& & \\
&Two similar shades close to each other (\textit{implicit})& & \\
\midrule
\multirow{4}{*}{State Change} &a small dark one \underline{traveling} southwest / 2 other dots \underline{following} it& \multirow{4}{*}{32.7\%} & \multirow{4}{*}{0.97} \\
&Do you have two dark med-size dots \underline{move slowly apart} as they \underline{drift} right? & & \\
&I have a large pale grey that \underline{moves down} but starts out \underline{curving} to the right and & & \\
&then \underline{takes a sharp turn} to the south east & & \\
\midrule
\multirow{3}{*}{Previous State} &I still see the larger gray one that \underline{was} next to it \underline{in the previous turn}. & \multirow{3}{*}{5.5\%}&  \multirow{3}{*}{0.79}\\
&I have the smaller dot that \underline{started out} below it to the left. & & \\
&\underline{Before it moves}, is there a lighter gray dot down and to the right of it?& & \\
\bottomrule
\end{tabular}
}
\caption{\label{tab:spatio_temporal_expressions}
Spatio-temporal expressions. Keywords (such as \textit{tense}, \textit{events} and \textit{motion verbs}) are underlined.
}
\end{table*}

\begin{table*}[th]
\centering \scalebox{0.83}{
\setlength\tabcolsep{9pt}
\setlength{\aboverulesep}{0pt}
\setlength{\belowrulesep}{0pt}
\setlength{\extrarowheight}{.4ex}
\begin{tabular}{cc|cc|cl}
\toprule
\multicolumn{2}{c|}{Degree Modifiers} & OCC & D-OCC & Examples (\# Keywords) & \multicolumn{1}{l}{Usage in D-OCC} \\
\midrule
\multirow{3}{*}{Scalar} & Diminishers & 9.2 & 8.9 & a bit, faintly, slightly (10) & \textbf{slightly} curves up \\
& Moderators & 1.3 & 0.9 & fairly, rather, somewhat (6) & \textbf{fairly} quickly \\
& Boosters & 9.8 & 6.1 & very, really, extraordinary (27) & \textbf{extremely} slowly \\
\midrule
\multirow{2}{*}{Totality} & Approximators & 10.2 & 6.4 & almost, maybe, probably (34) & \textbf{almost} collides with \\
& Maximizers & 4.3 & 4.2 & exactly, completely, definitely (37) & \textbf{perfectly} straight \\
\bottomrule
\end{tabular}
}
\caption{\label{tab:nuanced_expressions}
Average occurrences of degree modifiers per 100 utterances (estimated based on keywords).
}
\end{table*}

At the utterance level, we observed an extensive usage of \textit{spatio-temporal expressions} which are characteristic in dynamic environments. To study the frequency of such expressions, we manually annotated 100 dialogues in D-OCC with LST $\geq$ 2 (focusing on the more successful strategies).

Specifically, we detect whether each utterance contains 3 types of spatio-temporal expressions:\footnote{Note that a single utterance may contain none or multiple types of such expressions, and expressions of color, size or possession are not considered as spatio-temporal expressions.}

\begin{itemize}[topsep=0pt, itemsep=0pt, leftmargin=.2in, parsep=0pt]
    \item Reference to \textbf{current state} describes location of entities at the end of the current turn (i.e. timestep $t_k$ if the utterance is in turn $k$).
    \item Reference to \textbf{state change} describes temporal change of entity locations (i.e. movements).
    \item Reference to \textbf{previous state} describes entity locations at previous timestep $t$ (where $t < t_k$).
\end{itemize}

We show examples and estimated frequencies of spatio-temporal expressions in Table \ref{tab:spatio_temporal_expressions}. We also computed the agreement of our annotation based on 50 dialogues with 3 annotators, which we found to be reliable based on Cohen's $\kappa$ \citep{cohen1968weighted}.

Based on this result, we found that reference to \textit{state change} is the most widely used strategy, which could be simple as ``moves northwest'' or more complex as in Table \ref{tab:spatio_temporal_expressions}. Reference to \textit{previous state} is much less frequent compared to other types but still observed in many dialogues. Note that humans distinguish \textit{previous} and \textit{current} states in various ways, including temporal expressions (``was'', ``now''), motion verbs (``started out'', ``landed'') and implicit/default reasoning.

We also found that expressions are often \textit{nuanced} and \textit{pragmatic}, which are characteristic under continuous and partially-observable context \citep{udagawa2019natural}. 
Nuances are typically expressed by the \textit{degree modifiers} to convey subtle differences in location, movements, confidence, etc. Following \citet{paradis_2008}, we categorize them into 2 main types (and 5 subtypes): \textit{scalar modifiers} used for concepts in a range of scale (\textit{diminishers}, \textit{moderators}, \textit{boosters}) and \textit{totality modifiers} used for concepts with definite boundaries (\textit{approximators}, \textit{maximizers}). See Table \ref{tab:nuanced_expressions} for examples and the estimated occurrences of such modifiers in OCC and D-OCC.\footnote{Following the prior analysis in OCC, we manually curated keyword-based dictionaries of such modifiers (based on unigrams and bigrams) while removing polysemous words (such as \textit{little}, \textit{about}, \textit{too}, etc).} Based on these results, we can verify that there are comparable numbers of various degree modifiers in D-OCC as well, which are used effectively to cope with complex ambiguity and uncertainty.

\begin{figure}[t!]
\centering \small
\begin{tikzpicture}
\node[inner sep=0pt] (agent_0) at (0,0)
  {\includegraphics[width=0.46\columnwidth]{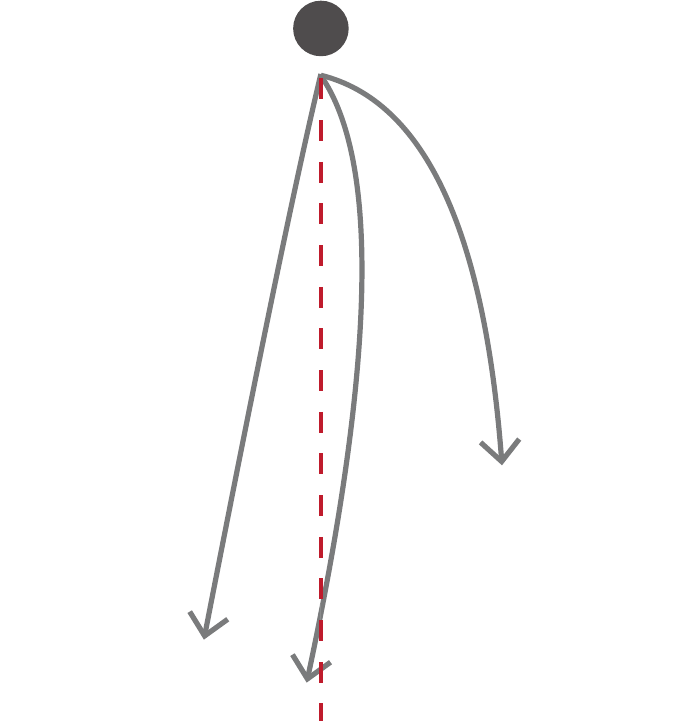}};
\node[inner sep=0pt] (agent_1) at (4,0)
  {\includegraphics[width=0.46\columnwidth]{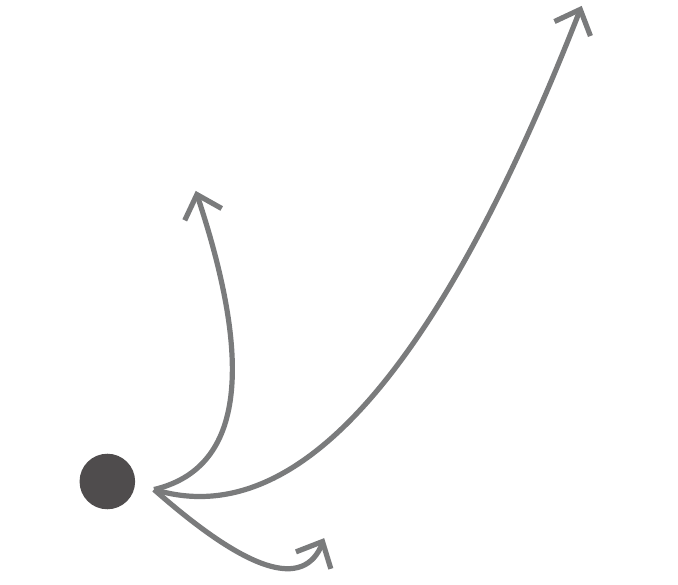}};
\node [below] at (0,-2) {straight down};
\node [below] at (4,-2) {right (and) then up};
\end{tikzpicture}
\caption{
Pragmatic expressions of movements.
}
\label{fig:pragmatic_movements}
\end{figure}

\begin{figure}[t!]
\centering \small
\begin{tikzpicture}
\node[inner sep=0pt] (fig_1) at (0,0)
  {\includegraphics[width=0.48\columnwidth]{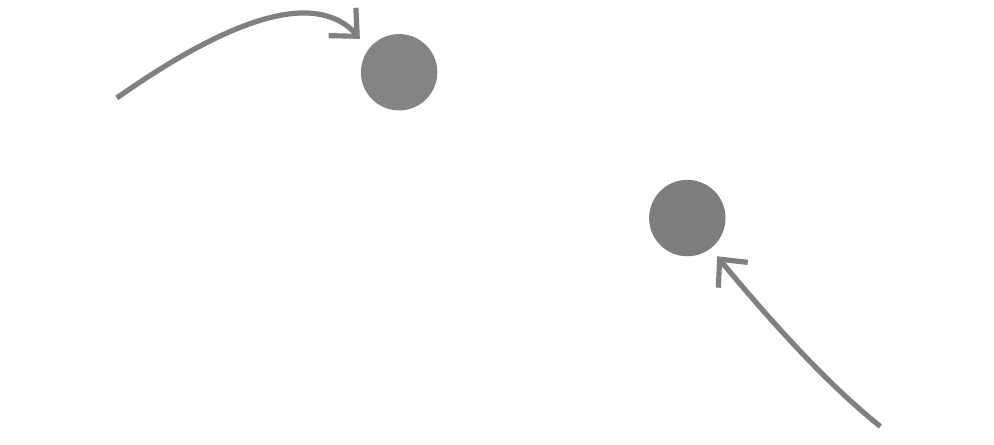}};
\node[inner sep=0pt] (fig_2) at (4,0)
  {\includegraphics[width=0.48\columnwidth]{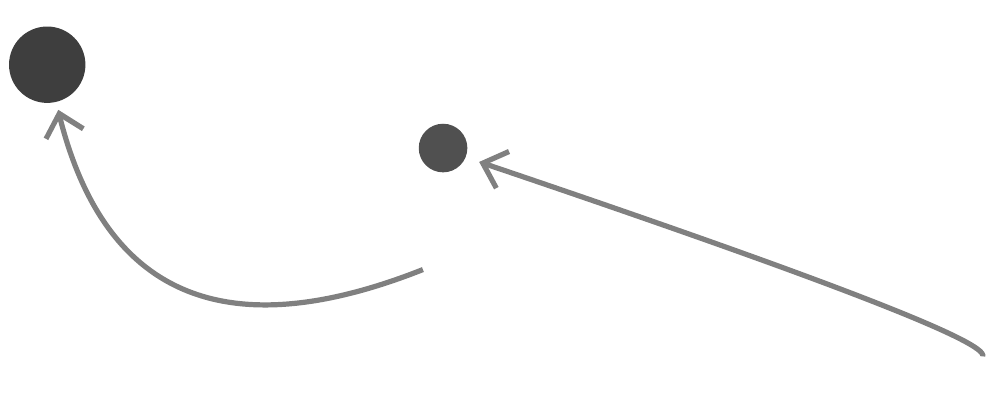}};
\node [below] at (0,-1) {\small toward each other};
\node [below] at (4.2,-1) {\small follow (behind)};
\node[inner sep=0pt] (fig_3) at (0,-3.3)
  {\includegraphics[width=0.46\columnwidth]{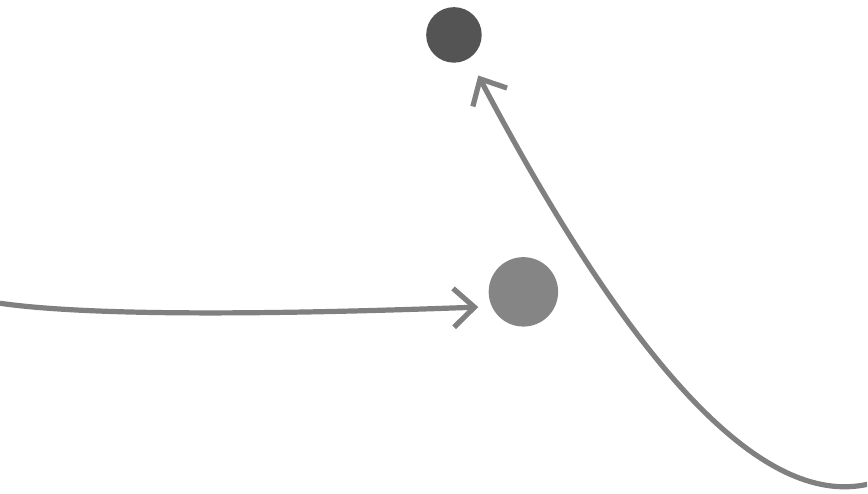}};
 \node [below] at (0,-4.8) {\small cross in front};
 \node[inner sep=0pt] (fig_4) at (4,-3.3)
  {\includegraphics[width=0.3\columnwidth]{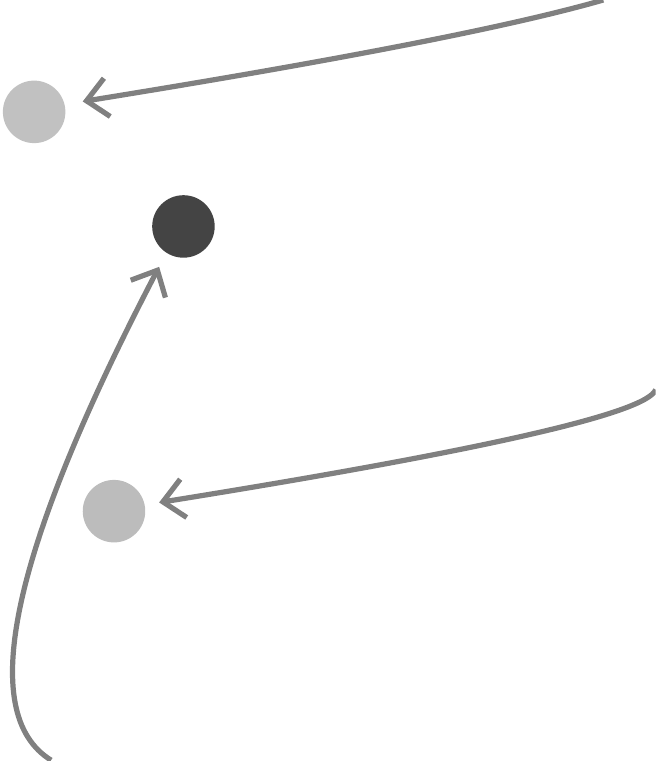}};
 \node [below] at (4,-4.8) {\small through (between)};
\end{tikzpicture}
\caption{
Expressions of multiple entity interactions.
}
\label{fig:entity_interactions}
\end{figure}
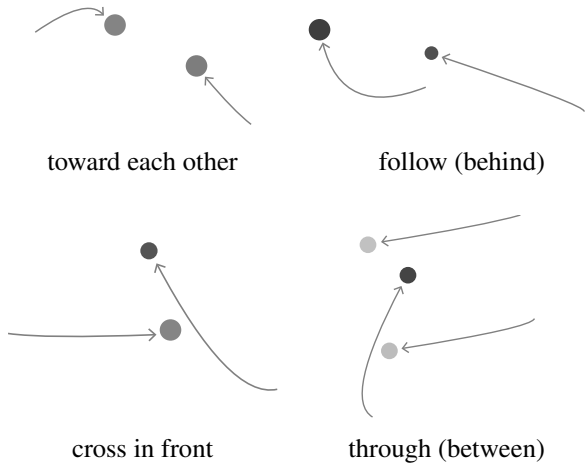

In Figure \ref{fig:pragmatic_movements}, we show examples of \textit{pragmatic} expressions which require pragmatic (non-literal) interpretations \citep{monroe2017colors}. For instance, trajectories of the expression ``straight down'' may not indicate vertical lines in the literal sense (e.g. could be curving or leaning to the left). Similarly, the expression of ``(moving) right and (then) up'' may be used for diverse movements ending up in various locations (e.g. even below the initial location!). While such expressions more or less deviate from literal semantics, they are pragmatically sufficient to convey the speaker's intention (i.e. identify the target among the distractors) \citep{Grice1975LogicAC}: alternatively, the speaker may need to choose different expressions for the same movement depending on the context (distractors).

We also show exemplary expressions of multiple entity interactions in Figure \ref{fig:entity_interactions}, which demonstrate interesting pragmaticality as well. For instance, ``toward each other'' may be used for trajectories moving in \textit{orthogonal} (rather than opposite) directions for the most of the time.

Overall, our analyses of spatio-temporal expressions reveal advanced language understanding and generation required in D-OCC, regardless of the task/lexical simplicity.

\begin{table*}[t!]
\centering \scalebox{0.83}{
\setlength\tabcolsep{8pt}
\setlength{\aboverulesep}{0pt}
\setlength{\belowrulesep}{0pt}
\setlength{\extrarowheight}{.4ex}
\newcommand{\intermidrule}{\cmidrule{2-8}}
\newcommand{\intermidmidrule}{\cmidrule{2-8}} 
\begin{tabular}{ccc|ccccc}
\toprule
\multirow{2}{*}{Dataset} & \multirow{2}{*}{Turn} & \raisebox{-1pt}{Previous} & \multicolumn{3}{c}{\raisebox{-1pt}{Success Rate (\%)}} & \raisebox{-1pt}{Utterances} & \raisebox{-1pt}{Tokens per} \\
 & & \raisebox{1pt}{Target} & \raisebox{1pt}{\#Shared=4} & \raisebox{1pt}{\#Shared=5} & \raisebox{1pt}{\#Shared=6} & \raisebox{1pt}{per Turn} & \raisebox{1pt}{Utterance} \\
\midrule
OCC & \nth{1} & - & 65.8 & 77.0 & 87.0 & 4.8 & 12.4 \\
\midrule
\multirow{3}{*}{D-OCC} & \nth{1} & - & 73.4 & 82.0 & 87.6 & 3.2 & 11.0 \\
 & \multirow{2}{*}{$\geq$\nth{2}} & \cmark & 95.4 & 97.0 & 97.8 & 2.3 & \phantom{0}5.9 \\
 & & \xmark & 81.7 & 88.4 & 91.6 & 3.5 & 11.7 \\
\bottomrule
\end{tabular}
}
\caption{\label{tab:turn_level_statistics}
Turn-level statistics of OCC and D-OCC. \,\cmark\,denotes cases where the previous target stays in common and \,\xmark\,denotes it left at least one agent's view. Note that \# shared entities are 4, 5 or 6 at selection timesteps (\cref{subsec:sequential_collaborative_reference_task}).
}
\end{table*}

\begin{table*}[t!]
\centering \scalebox{0.83}{
\setlength\tabcolsep{8pt}
\begin{tabular}{clc}
\toprule
\multicolumn{1}{c}{Previous Target} & \multicolumn{1}{c}{Examples} & Freq. \\
\midrule
\multirow{2}{*}{Stay (\cmark)} & I still see the same dot / I still have all three dots from the line before & \multirow{2}{*}{36.8\%} \\
& Left my screen, but may have come back traveling left to right? &  \\
\midrule
\multirow{2}{*}{Leave (\xmark)} & I lost the last one / I lost the light one but still see the darker one that was on its left. & \multirow{2}{*}{63.2\%} \\
& both are gone for me / similar size black dot that barely moves? (\textit{implicit}) \\
\bottomrule
\end{tabular}
}
\caption{\label{tab:turn_level_utterances}
Comparison of utterances when the previous target stays in common (\cmark) or not (\xmark).
}
\end{table*}

\subsection{Turn-Level Strategies}
\label{subsec:turn_level_strategies}

Finally, we study and compare human strategies at different timesteps (in different turns). Table \ref{tab:turn_level_statistics} shows detailed statistics of the dataset in the initial turn and later turns, where \textit{creation} and \textit{maintenance} of common ground are required, respectively. Note that we also distinguish later turns based on whether the previous selection (i.e. \textit{previous target}) stays in common (\cmark) or leaves at least one agent's view (\xmark): former cases can \textit{retain} the same common ground but the latter cases require an \textit{update} of common ground.

First, if we focus on the \nth{1} turn, we can verify that success rates are consistently higher in D-OCC than OCC, especially in difficult cases when the number of shared entities is smaller. This indicates that humans can create common ground more accurately by leveraging dynamic information (e.g. entity movements) unavailable in OCC.

In later turns, we found that human performance is near perfect with shorter dialogues in \cmark\,cases (when the previous target stays in common). This is natural because they can simply retain common ground and repeat the same selection. Notably, human performance is consistently higher than the \nth{1} turn \textit{even in} \xmark\,cases (when the previous target is no longer in common), which verifies that humans can leverage previous common ground to \textit{update} common ground more reliably as well.

We show example utterances of \cmark\,and \xmark\,cases in Table \ref{tab:turn_level_utterances}. Note that the previous target may temporarily leave the view and come back in \cmark\,cases, which occasionally makes even \textit{retainment} of the same common ground non-trivial. In \xmark\,cases, humans either inform about the lost entities explicitly or \textit{implicitly}, e.g. by ignoring old entities and starting to focus on the new ones.

\section{Experiments}
\label{sec:experiments}

Finally, we conduct extensive experiments to assess our baseline model's capability of common grounding in dynamic environments.

\subsection{Evaluation}
\label{subsec:evaluation}

To study the model's capability from various aspects, we design 3 (sub)tasks based on D-OCC.

First, we evaluate the model's ability of \textit{recognizing} common ground based on the \textbf{target selection task}, originally proposed for OCC. This is an important subtask of (sequential) collaborative reference, where the model is given one player's observation and the (ground-truth) dialogue history to predict which target was selected by the player. Since there can be multiple selections in D-OCC, the model makes predictions at the end of each turn $k$ (at timestep $t_k$). The number of entities observable at $t_k$ is fixed at 7 for both OCC and D-OCC (\cref{subsec:sequential_collaborative_reference_task}), so this is a simple classification task evaluated based on accuracy.

Secondly, we estimate the model's ability of \textit{creating} and \textit{maintaining} common ground based on the \textbf{selfplay dialogue task}, where each model plays the full sequential collaborative reference task against an identical copy of itself. While this evaluation has the advantage of being scalable and automatic, succeeding on this setting is only \textit{necessary} for human-level common grounding and not \textit{sufficient}, since the model may only be able to coordinate with itself (and not with real humans).

Thirdly, we conduct \textbf{human evaluation} to test the model's ability of playing sequential collaborative reference against real human workers on AMT. Due to the high cost of this evaluation, we only focus on the top 3 variants of our baseline ranked by avg. LST in the selfplay dialogue task.

\subsection{Model Architecture}
\label{subsec:model}

\begin{figure*}[tb!]
\centering
\includegraphics[width=\textwidth]{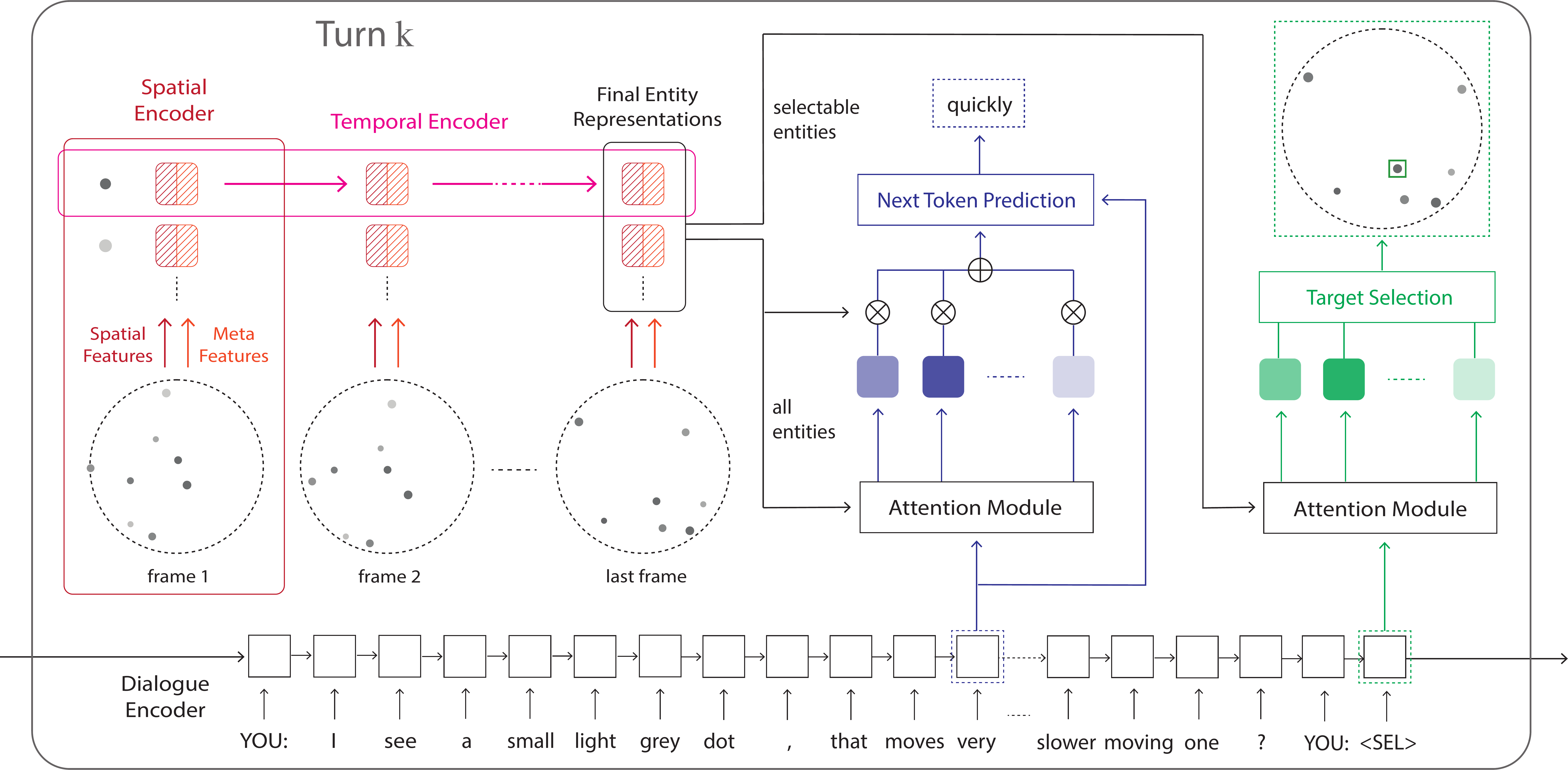}
\caption{Our baseline model architecture. Information flow in turn $k$ is illustrated.
When generating model utterances (in selfplay dialogue and human evaluation), we sample next tokens with the temperature set to 0.25.
}
\label{fig:model_architecture}
\end{figure*}

For a fair comparison with prior work, we implement our baseline model following the OCC models in \citet{udagawa2020annotated}. The overall model architecture is shown in Figure \ref{fig:model_architecture}.

To encode the dialogue tokens throughout the turns, we use a unidirectional GRU \citep{cho2014properties}. To encode the observation during turn $k$, we first split the animation of entity movements into 10 frames and the agent view shift into 5 frames. Then, we process each observation frame based on the \textit{spatial} encoder, followed by the \textit{temporal} encoder to integrate these outputs.

The spatial encoder is used to extract \textit{spatial features} and \textit{meta features} from each observation frame. Spatial features represent the spatial attributes of each entity (color, size and location in the frame), which are encoded using an MLP and a relation network \citep{santoro2017simple}. The relation network is used to represent the spatial attributes relative to a subset of entities $\tilde{E} \subset E$, which could be \textit{all entities} observable in turn $k$ ($E_{all}$) or \textit{selectable entities} visible at $t_k$ ($E_{sel}$). Hence, the spatial features of $e_i$ are computed as:
\begin{equation}\label{eqt:spatial_features}
\mathrm{MLP}(\mathbf{e}_i) \odot \sum_{\substack{e_j \in \tilde E, \\ j \neq i }} \mathrm{MLP}(\mathbf{e}_i - \mathbf{e}_j)
\end{equation}
where  $\mathbf{e}_i$ is the vector representation of entity $e_i$ and $\odot$ is the vector concatenation.\footnote{To be precise, $\mathbf{e}_i$ is a 4-dimensional vector representing color, size, and 2-D location. If the entity is not observable in the frame, we use the default value of (0, 0) for the location.}

Meta features are binary information of each entity representing whether (or not) the entity (i) is visible in the frame, (ii) is visible at timestep $t_{k}$, (iii) was visible at timestep $t_{k-1}$, and (iv) was selected in the previous turn (i.e. is the \textit{previous target}). Meta features are also encoded using an MLP, and we take the sum of spatial/meta features as the (entity-level) output of the spatial encoder.

Finally, we use the temporal encoder based on a GRU to encode the outputs of the spatial encoder. The final state of the temporal encoder is considered as the final representation of each entity.

Based on the outputs of these encoders, we use two attention modules (based on MLPs) to compute attention scores for each entity. The first attention module is used to weight the final representations of all entities $E_{all}$ conditioned on the current dialogue state: then, the weighted sum of $E_{all}$ is concatenated with the dialogue state to predict the next dialogue token \citep{xu2015show}. The second module is used to predict the target entity, where we simply take the (soft)max of attention scores for the selectable entities $E_{sel}$ in turn $k$.

Note that there are only two main differences between our baseline and the best OCC model (TSEL-REF-DIAL) from \citet{udagawa2020annotated}: first, in TSEL-REF-DIAL, the final representation of each entity is its \textit{spatial features}, i.e. the meta features and temporal encoder are not used (which are only meaningful in D-OCC). Second, TSEL-REF-DIAL is also trained on the \textit{reference resolution} task (using an additional attention module), which is only available in OCC. Due to this architectural similarity, we can virtually \textit{pretrain} our model on OCC by initializing the shared model parameters based on TSEL-REF-DIAL and then fine-tune the whole model on D-OCC.\footnote{For pretraining, we retrained TSEL-REF-DIAL with the shared word embedding for OCC and D-OCC.}

\subsection{Experiment Setup}
\label{subsec:experiment_setup}

All modules of our baseline (MLPs and GRUs) are single layered with 256 hidden units, except for the attention modules which are 2-layered. Dropout rate of 0.5 is applied at each layer during training, and we use the Adam optimizer \citep{Kingma2015AdamAM} with the initial learning rate set to 0.001. After manual tuning on the validation set, we weight the losses from next token prediction and target selection with the ratio of 2:1.

In terms of data splits, we use 500 dialogues with LST $\geq$ 2 for testing target selection, another 500 for validation and the rest for training.\footnote{We ensured no overlaps in terms of the \textit{environments} across data splits.} Note that we use all unsuccessful turns (where the players failed to agree upon the same entity) as well, assuming they are still based on valid strategies. For selfplay dialogue and human evaluation, we collect 2,000 and 200 dialogues in unseen environments, respectively. Each experiment is repeated 5 times with different random seeds (including data splits), except for human evaluation.

Finally, we conduct extensive ablations to study the effect of various model architectures, including \textit{pretraining}, spatial attributes (\textit{color}, \textit{size} and \textit{location}) and the meta feature (\textit{previous target}). In addition, we also ablate the \textit{dynamic} information of the observation by only using the last frame in each turn as the input for the temporal encoder.

\subsection{Results}
\label{subsec:results}

\begin{table}[ht]
\centering \scalebox{0.83}{
\setlength\tabcolsep{6pt}
\begin{tabular}{lccc}
\toprule
\multirow{2}{*}[-0.8ex]{Model}& \multicolumn{3}{c}{Turn / Previous Target} \\
\cmidrule{2-4}
 & \nth{1} / - & $\geq$\nth{2} / \cmark & $\geq$\nth{2} / \xmark \\
\midrule
Baseline & \textbf{76.4{\scriptsize $\pm$1.7}} & 96.6{\scriptsize $\pm$0.3}\phantom{$^*$} & 67.4{\scriptsize $\pm$0.5}\phantom{$^*$} \\
\phantom{F}-- pretraining & 74.6{\scriptsize $\pm$2.7} & 96.3{\scriptsize $\pm$0.7}\phantom{$^*$} & 66.9{\scriptsize $\pm$1.1}\phantom{$^*$} \\
\phantom{F}-- color & 56.3{\scriptsize $\pm$2.0} & 95.7{\scriptsize $\pm$0.6}\phantom{$^*$} & 50.5{\scriptsize $\pm$1.4}\phantom{$^*$} \\
\phantom{F}-- size & 58.4{\scriptsize $\pm$1.3} & 95.7{\scriptsize $\pm$0.9}\phantom{$^*$} & 52.2{\scriptsize $\pm$0.5}\phantom{$^*$} \\
\phantom{F}-- location & 74.4{\scriptsize $\pm$1.5} & 96.1{\scriptsize $\pm$0.9}\phantom{$^*$} & 67.3{\scriptsize $\pm$0.7}\phantom{$^*$} \\
\phantom{F}-- previous target & 76.1{\scriptsize $\pm$1.7} & 83.3{\scriptsize $\pm$1.1}$^*$ & \textbf{67.8{\scriptsize $\pm$0.6}}$^*$ \\
\phantom{F}-- dynamics & 75.1{\scriptsize $\pm$2.3} & \textbf{96.7{\scriptsize $\pm$1.0}}\phantom{$^*$} & 67.0{\scriptsize $\pm$0.7}\phantom{$^*$} \\
\midrule
Human & 97.0{\scriptsize $\pm$1.1} & 98.2{\scriptsize $\pm$0.5}$^*$ & 95.8{\scriptsize $\pm$2.0}$^*$ \\
\bottomrule
\end{tabular}
}
\caption{\label{tab:model_target_selection}
Results for the \textbf{target selection task} ($^*$ denotes cases where the correct previous targets were not provided during prediction).
}
\end{table}

\begin{table*}[th!]
\scalebox{0.77}{
\centering
\setlength\tabcolsep{6pt}
\setlength{\aboverulesep}{0pt}
\setlength{\belowrulesep}{0pt}
\setlength{\extrarowheight}{.1ex}
\newcommand{\intermidrule}{\cmidrule{2-8}}
\newcommand{\intermidmidrule}{\cmidrule{2-8}} 
\begin{tabular}{l|ccc|cccc|cc}
\toprule
\multicolumn{1}{c}{} & & & \multicolumn{1}{c}{} & \multicolumn{4}{|c}{\textbf{Seflplay Dialogue}} & \multicolumn{2}{|c}{\textbf{Human Evaluation}} \\
\midrule
\midrule
\multirow{2}{*}{Model}& \multirow{2}{*}{Dataset} & \multirow{2}{*}{Turn} & \raisebox{-0.5pt}{Previous} & \multicolumn{3}{c}{\raisebox{-0.5pt}{Success Rate (\%)}} & \raisebox{-0.5pt}{Avg.} & \raisebox{-0.5pt}{\phantom{}Success\phantom{(}} & \raisebox{-0.5pt}{Avg.} \\
 & & & \raisebox{0.5pt}{Target} & \raisebox{0.5pt}{\#Shared=4} & \raisebox{0.5pt}{\#Shared=5} & \raisebox{0.5pt}{\#Shared=6} & \raisebox{0.5pt}{LST} & \raisebox{0.5pt}{Rate (\%)} & \raisebox{0.5pt}{LST} \\
\midrule
\multirow{3}{*}{Baseline} & \multirow{3}{*}{D-OCC} & \nth{1} & - & 46.8{\scriptsize $\pm$1.8} & 63.8{\scriptsize $\pm$1.8} & 80.2{\scriptsize $\pm$2.3} & \multirow{3}{*}{1.94{\scriptsize $\pm$0.09}} & \textbf{44.5} & \multirow{3}{*}{\textbf{1.00}} \\
 & & \multirow{2}{*}{$\geq$\nth{2}} & \cmark & 99.4{\scriptsize $\pm$0.3} & 99.7{\scriptsize $\pm$0.2} & 99.6{\scriptsize $\pm$0.2} & & 81.9 & \\
 & & & \xmark & \textbf{48.5{\scriptsize $\pm$2.2}} & \textbf{64.6{\scriptsize $\pm$2.8}} & \textbf{81.5{\scriptsize $\pm$1.5}} & & \textbf{44.4} & \\
\midrule
\multirow{3}{*}{\phantom{F}-- pretraining} & \multirow{3}{*}{D-OCC} & \nth{1} & - & 39.4{\scriptsize $\pm$1.0} & 53.5{\scriptsize $\pm$0.8} & 73.7{\scriptsize $\pm$1.8} & \multirow{3}{*}{1.35{\scriptsize $\pm$0.09}} & \multirow{3}{*}{N/A} & \multirow{3}{*}{N/A} \\
 & & \multirow{2}{*}{$\geq$\nth{2}} & \cmark & 98.6{\scriptsize $\pm$2.4} & 98.8{\scriptsize $\pm$1.8} & 99.4{\scriptsize $\pm$1.0} & & & \\
 & & & \xmark & 30.3{\scriptsize $\pm$5.7} & 42.1{\scriptsize $\pm$6.3} & 65.4{\scriptsize $\pm$4.9} & & & \\
\midrule
\multirow{3}{*}{\phantom{F}-- color} & \multirow{3}{*}{D-OCC} & \nth{1} & - & 36.3{\scriptsize $\pm$2.0} & 54.6{\scriptsize $\pm$2.3} & 72.9{\scriptsize $\pm$1.5} & \multirow{3}{*}{1.50{\scriptsize $\pm$0.10}} & \multirow{3}{*}{N/A} & \multirow{3}{*}{N/A} \\
 & & \multirow{2}{*}{$\geq$\nth{2}} & \cmark & 99.7{\scriptsize $\pm$0.1} & 99.7{\scriptsize $\pm$0.0} & 99.6{\scriptsize $\pm$0.1} & & & \\
 & & & \xmark & 42.1{\scriptsize $\pm$3.5} & 56.7{\scriptsize $\pm$4.2} & 72.4{\scriptsize $\pm$4.6} & & & \\
\midrule
\multirow{3}{*}{\phantom{F}-- size} & \multirow{3}{*}{D-OCC} & \nth{1} & - & 41.5{\scriptsize $\pm$0.8} & 58.0{\scriptsize $\pm$0.9} & 75.2{\scriptsize $\pm$1.3} & \multirow{3}{*}{1.58{\scriptsize $\pm$0.07}} & \multirow{3}{*}{N/A} & \multirow{3}{*}{N/A} \\
 & & \multirow{2}{*}{$\geq$\nth{2}} & \cmark & 99.8{\scriptsize $\pm$0.1} & 99.7{\scriptsize $\pm$0.1} & 99.8{\scriptsize $\pm$0.2} & & & \\
 & & & \xmark & 39.6{\scriptsize $\pm$3.5} & 55.3{\scriptsize $\pm$3.6} & 69.9{\scriptsize $\pm$1.5} & & & \\
\midrule
\multirow{3}{*}{\phantom{F}-- location} & \multirow{3}{*}{D-OCC} & \nth{1} & - & 45.7{\scriptsize $\pm$1.9} & 60.4{\scriptsize $\pm$1.6} & 77.7{\scriptsize $\pm$1.7} & \multirow{3}{*}{1.68{\scriptsize $\pm$0.09}} & 40.0 & \multirow{3}{*}{0.81} \\
 & & \multirow{2}{*}{$\geq$\nth{2}} & \cmark & 99.8{\scriptsize $\pm$0.1} & 99.7{\scriptsize $\pm$0.0} & 99.7{\scriptsize $\pm$0.1} & & \textbf{91.8} & \\
 & & & \xmark & 40.8{\scriptsize $\pm$3.6} & 54.6{\scriptsize $\pm$2.5} & 73.9{\scriptsize $\pm$4.2} & & 36.3 & \\
\midrule
\multirow{3}{*}{\phantom{F}-- previous target} & \multirow{3}{*}{D-OCC} & \nth{1} & - & \textbf{49.2{\scriptsize $\pm$1.3}} & 64.0{\scriptsize $\pm$1.8} & 82.2{\scriptsize $\pm$2.0} & \multirow{3}{*}{1.45{\scriptsize $\pm$0.05}} & \multirow{3}{*}{N/A} & \multirow{3}{*}{N/A} \\
 & & \multirow{2}{*}{$\geq$\nth{2}} & \cmark & 85.8{\scriptsize $\pm$2.7} & 87.5{\scriptsize $\pm$1.6} & 91.2{\scriptsize $\pm$1.3} & & & \\
 & & & \xmark & 29.2{\scriptsize $\pm$1.5} & 41.9{\scriptsize $\pm$1.9} & 64.5{\scriptsize $\pm$1.0} & & & \\
\midrule
\multirow{3}{*}{\phantom{F}-- dynamics} & \multirow{3}{*}{D-OCC} & \nth{1} & - & \textbf{49.2{\scriptsize $\pm$2.2}} & \textbf{65.8{\scriptsize $\pm$1.3}} & \textbf{83.3{\scriptsize $\pm$1.9}} & \multirow{3}{*}{\textbf{2.02{\scriptsize $\pm$0.07}}} & 37.0 & \multirow{3}{*}{0.79} \\
 & & \multirow{2}{*}{$\geq$\nth{2}} & \cmark & \textbf{99.9{\scriptsize $\pm$0.1}} & \textbf{99.9{\scriptsize $\pm$0.1}} & \textbf{99.8{\scriptsize $\pm$0.1}} & & 86.8 & \\
 & & & \xmark & 48.3{\scriptsize $\pm$2.2} & 63.5{\scriptsize $\pm$2.8} & 81.1{\scriptsize $\pm$2.1} & & 39.2 & \\
\midrule
\multirow{2}{*}{TSEL-REF-DIAL} & D-OCC & \nth{1} & - & 41.0{\scriptsize $\pm$1.2} & 58.7{\scriptsize $\pm$1.1} & 76.0{\scriptsize $\pm$1.8} & \multirow{2}{*}{-} & \multirow{2}{*}{N/A} & \multirow{2}{*}{-} \\
& OCC & \nth{1} & - & 45.9{\scriptsize $\pm$1.6} & 62.7{\scriptsize $\pm$2.2} & 79.7{\scriptsize $\pm$1.0} & & & \\
\midrule
\midrule
\multirow{3}{*}{Human} & \multirow{3}{*}{D-OCC} & \nth{1} & - & 73.4 & 82.0 & 87.6 & \multirow{3}{*}{3.31} & 80.5 & \multirow{3}{*}{3.31} \\
 & & \multirow{2}{*}{$\geq$\nth{2}} & \cmark & 95.4 & 97.0 & 97.8 & & 96.7 & \\
 & & & \xmark & 81.7 & 88.4 & 91.6 & & 86.6 & \\
\bottomrule
\end{tabular}
}
\caption{\label{tab:model_full_dialogue}
Results for the sequential collaborative reference task (\textbf{selfplay dialogue} and \textbf{human evaluation}). Human performance is estimated based on the overall average of the crowd workers (c.f. Table \ref{tab:onecommon_statistics} and \ref{tab:turn_level_statistics}).
}
\end{table*}

We show the results for target selection in Table \ref{tab:model_target_selection}. The human performance is estimated by 3 annotators based on 50 dialogues with LST $\geq$ 2.

Based on these results, we can verify that all ablations hurt the performance of our baseline in some way. Pretraining on OCC is generally effective, and all spatial attributes contribute to the overall performance (especially color and size). When the meta feature of the correct previous target is available, all models perform remarkably well in \cmark\,cases (previous target stays in common), which is natural since humans often repeated the same selection. Finally, dynamic information also contributes to the baseline performance, despite the effect being rather marginal.

However, there is huge room left for improvement in the \nth{1} turn and even more so in \xmark\,cases (previous target no longer in common). These results indicate that recognizing the \textit{creation} of common ground is still difficult, and recognizing how they are \textit{updated} (rather than \textit{retained}) remains even more challenging for the current baseline.

Next, we show the results for selfplay dialogue and human evaluation in Table \ref{tab:model_full_dialogue}. We also include the results of TSEL-REF-DIAL (trained on OCC \textit{without} fine-tuning on D-OCC) as a reference.\footnote{When testing TSEL-REF-DIAL on D-OCC, we used the spatial features of the last observation frame as the input.}

In selfplay dialogue, we can verify that the baseline model performs reasonably well, outperforming TSEL-REF-DIAL in the \nth{1} turn of D-OCC (as well as OCC). However, it is worth noting that TSEL-REF-DIAL may be suffering from a minor covariate shift in D-OCC (c.f. \cref{subsec:sequential_collaborative_reference_task}), and without pretraining, our baseline still underperforms this best OCC model. We also found that all ablations of spatial attributes hurt performance, while the locational attributes became more critical in the full dialogue task. The meta feature of the previous target (selected by the model) is also critical, as the models seem to be relying heavily on this feature to both retain and update the target.

However, we found that ablation of dynamic information does not degrade (actually improves) performance in selfplay dialogue. This indicates that the last frame of each turn (\textit{current state}) is sufficient for the baseline to coordinate with itself, and it is unlikely to be leveraging sophisticated temporal information (\textit{state change} or \textit{previous state}) like the human strategies seen in \cref{subsec:spatio_temporal_expressions}. Also, while the models perform near perfectly in \cmark\,cases, the success rates drop or do not improve significantly in \xmark\,cases (compared to the \nth{1} turn). This shows that current models can \textit{retain} the same common ground easily but struggle in \textit{updating} them using the previous common ground, unlike the human strategies seen in \cref{subsec:turn_level_strategies}.

Finally, in human evaluation, we could verify that our baseline performs the best of the top 3 models in the selfplay dialogue task, but the success rates were much lower than observed in selfplay. This indicates that current models may not be using natural language in the same way humans use it (i.e. are not properly \textit{grounded}, \citealp{bender-koller-2020-climbing}), although they do become closer to it when all the features are available.\footnote{At the \textit{superficial} level, all models could generate fluent utterances and complete the task with minimal confusion.}

To summarize, our results in sequential collaborative reference show that the current baseline can leverage all spatial features and retain the same common ground, especially when provided explicitly as the meta feature. However, it may not be using temporal information effectively, and the creation and update of common ground still remain challenging in the dynamic environments, especially when conversing with real humans.

\section{Discussion and Conclusion}
\label{sec:discussion_and_conclusion}

In this work, we proposed a novel dialogue task to study the ability of creating, retaining and updating common ground in dynamic environments. The introduced dynamics are fully controllable in our setting to maximize diversity, minimize biases and enable reliable evaluation and analysis. Based on our dataset analyses and experiments, we demonstrated the advanced strategies of common grounding required and the open room for improvement in our newly developed Dynamic-OneCommon Corpus (D-OCC).

In future work, we plan to utilize and enrich this dataset in several ways. For instance, we can conduct various \textit{causal analysis}, e.g. by changing certain feature of entities (such as movement) and studying the differences in model behavior, which is essential yet difficult to conduct in many existing datasets (c.f. \cref{sec:related_work}). Another promising direction is to add fine-grained annotation of \textit{reference resolution} \citep{udagawa2020annotated}, as (partially) illustrated in Figure \ref{fig:first_example}. We can also annotate \textit{spatio-temporal expressions}, e.g. by following the procedure in \citet{udagawa-etal-2020-linguistic}. Such annotations would allow us to gain deeper understandings of the \textit{intermediate} process of common grounding: for instance, we can study whether the developed models recognize and use the spatio-temporal expressions appropriately and consistently in a \textit{human-like} way (i.e. not only imitate at the superficial level, as observed in \cref{subsec:results}).

In order to improve the model performance, we're considering several approaches. One approach is to make the model learn from task success (and failure) through \textit{reinforcement learning}. Due to the symmetric agent roles in our task, this is straightforward to conduct through selfplay \citep{lewis-etal-2017-deal,DBLP:conf/icml/YaratsL18}, and we can expect the models to avoid ineffective strategies like underspecification and premature guessing. We also expect the incorporation of \textit{pragmatic reasoning} to be a fruitful area of future research. One representative approach is the \textit{Rational Speech Act} (\textit{RSA}) framework \citep{Goodman2016PragmaticLI}, which has been applied in both continuous \citep{monroe2017colors} and partially-observable domains \citep{Hawkins2021TheDO}. However, application in \textit{dynamic} domains would involve additional complexities that need to be taken into account, such as the dependencies on previous common ground. Finally, we're planning to study wider variety of model architectures and pretraining datasets, including video-processing methods \citep{Carreira2017QuoVA,wang2018non}, vision-language grounding models \citep{lu2019vilbert,le-etal-2020-bist} and large-scale, open domain datasets \citep{krishna2017visual,sharma-etal-2018-conceptual}. Note that the entity-level representation of the observation (required in our baseline) can be obtained from raw video features, e.g. by utilizing the object trackers \citep{tracktor_2019_ICCV,wang2020towards}.

Finally, we'd like to discuss the main limitation of our current work, namely the \textit{ecological validity} \citep{de2020towards} of D-OCC. Since we focused on the simplest task setting under \textit{continuous}, \textit{partially-observable} and \textit{dynamic} context, direct application of our work in realistic settings may not be straightforward. However, the generic strategies required in our setting are fundamental in many real-world applications. For an illustration, imagine a navigation task in a \textit{dynamic} environment, such as finding a lost child in an urban city. Since the \textit{target entity} (the child) may not stay in one place, routing directions can no longer be fixed and need to be \textit{updated} accordingly (as in ``now head more to the west'' or ``go back to the previous block''). Furthermore, the \textit{landmark entities} may not be stationary either and could be ephemeral (as in ``following the group of travelers'' or ``in the middle of the crowd''). Lastly, if the child is not conspicuous with confusable distractors (e.g. with many pedestrians around), the descriptions need to be precise and distinguishing (as in ``wearing \textit{a little} darker shirt'' or ``walking \textit{right} towards the station'').

In order to study such (nuanced and pragmatic) spatio-temporal expressions and references to previous common ground, we expect D-OCC to be an essential proving ground. In addition, our sequential collaborative reference task is defined generally (c.f. \cref{subsec:sequential_collaborative_reference_task}), so we can easily scale up the task complexity to study the desired dynamics under consideration: the exploration of different, potentially more complex dynamics is an important research area left as future work.

Overall, we expect our task design, resource and analyses to be fundamental for developing dialogue systems that can both create and maintain common ground in dynamic environments.

\section*{Acknowledgments}

We are grateful to our action editor, Michel Galley, and the three anonymous reviewers for their
valuable suggestions that helped improve this paper. We also thank Saku Sugawara and Taichi Iki for their constructive feedback on earlier versions of this paper. This work was supported by JSPS KAKENHI Grant Number 21H03502.

\bibliography{tacl2021}
\bibliographystyle{acl_natbib}

\end{document}